%% file: main.tex
\documentclass{article}
\usepackage[preprint]{spconf}
\usepackage{amsmath,graphicx}
\usepackage{amsfonts}
\usepackage{amssymb}
\include{macros}
\usepackage{multirow}
\usepackage{siunitx}

\copyrightnotice{\copyright\ IEEE 2024}
\toappear{To appear in {\it Proc.\ ICASSP2024,
   April 14-19, 2024, Seoul, Korea}}

\title{Keep decoding parallel with Effective Knowledge Distillation from Language Models to End-to-end Speech Recognisers}

\name{Michael Hentschel$^{1,\star}$\thanks{$^{\star}$equal contribution}, Yuta Nishikawa$^{2,\star}$, Tatsuya Komatsu$^{3}$, Yusuke Fujita$^{3}$}
\address{$^{1}$LINE WORKS Corporation, Japan \quad $^{2}$Nara Institute of Science and Technology, Japan \\ $^{3}$LINE Corporation, Japan \quad}
\begin{document}
\ninept
\maketitle
\begin{abstract}
This study presents a novel approach for knowledge distillation (KD) from a BERT teacher model to an automatic speech recognition (ASR) model using intermediate layers. To distil the teacher's knowledge, we use an attention decoder that learns from BERT's token probabilities. Our method shows that language model (LM) information can be more effectively distilled into an ASR model using both the intermediate layers and the final layer. By using the intermediate layers as distillation target, we can more effectively distil LM knowledge into the lower network layers. Using our method, we achieve better recognition accuracy than with shallow fusion of an external LM, allowing us to maintain fast parallel decoding. Experiments on the LibriSpeech dataset demonstrate the effectiveness of our approach in enhancing greedy decoding with connectionist temporal classification (CTC).
\end{abstract}

\section{Introduction}
\label{sec:intro}

End-to-end automatic speech recognition (E2E-ASR) systems are reported to learn an internal language model (LM) \cite{meng2021internal, meng2021internaltraining}.
Their training requires a large amount of speech and its transcription text as training data.
If the amount of training data is insufficient, the ASR output may incorrectly contain linguistically meaningless token sequences for unknown words or unseen input sequences.
In order to cope with these errors, it is necessary to combine an external LM, learned from large-scale data, with the E2E-ASR system in shallow fusion \cite{Chorowski2017, kannan18_icassp} during inference.
Especially in the case of a non-autoregressive ASR model, high inference speed and high throughput are sacrificed by autoregressive decoding in this shallow fusion.
Although autoregressive E2E-ASR models such as recurrent neural network transducers (RNN-T) \cite{Graves2012Sequence,graves2013speech} are popular, in batch processing applications that require the processing of many requests in parallel a non-autoregressive ASR model is preferred for its higher throughput.
Hence, we aim to strengthen the internal LM of a connectionist temporal classification (CTC) \cite{graves2006connectionist} type  non-autoregressive ASR model (CTC-ASR) in this study.
Our proposed method allows to realise a CTC-ASR model with high recognition accuracy during non-autoregressive decoding without requiring shallow fusion with an external LM.

Training the internal LM of an {E2E-ASR} with only text data is not a simple task. Several methods for knowledge distillation (KD) \cite{hinton2015distilling} from an external LM trained on a large-scale text corpus have been proposed in previous studies.
BERT \cite{devlin2019bert} is often adopted as an external LM used as a teacher model for KD.
\cite{futami2020distilling} proposed to use the posterior distribution output from BERT as a soft label to improve an autoregressive ASR model.
Further studies proposed to design an auxiliary task as a regression problem to distill the hidden state representations of BERT in combination with non-autoregressive ASR \cite{bai2021fast} and RNN-T \cite{kubo2022knowledge}.
Higuchi et al. \cite{higuchi-etal-2022-bert} suggested a method for transferring BERT language information using an iterative refinement scheme.

In general, KD trains a student model to produce the same output as a teacher model for a given input.
Recently, a more efficient way for KD to transfer knowledge from a teacher model to a student model was proposed.
Several studies in the field of natural language processing showed that the use of intermediate loss functions was effective when distilling BERT's knowledge into student models \cite{jiao-etal-2020-tinybert,sun-etal-2020-mobilebert,hentschel2021making}.
In the case of CTC-ASR, \cite{yoon2022interkd} used an intermediate layer loss function when distilling a teacher E2E-ASR model into a student E2E-ASR model.

This work proposes an efficient KD method using an intermediate layer loss function for non-autoregressive CTC-ASR models.
In the proposed method, KD is applied via an auxiliary attention decoder (AED) to the intermediate layer  (interAED-KD) \cite{komatsu2023interdecoder} in addition to KD on the final layer \cite{deng2022improving,futami2020distilling}.
As with KD on the final layer, the auxiliary AED in the intermediate layer utilises BERT's posterior distribution as a teacher soft label.
In contrast to using the auxiliary loss function only on the final encoder layer, the auxiliary AED loss function on the intermediate layer can affect deeper layers of the encoder, enabling a more efficient KD.
Unlike in conventional KD \cite{futami2020distilling}, that makes inference with AEDs,
in this study, inference is performed by CTC decoding. An AED is used only as an auxiliary loss function during training. We show that even in the case of CTC decoding, KD using the AED auxiliary loss on the final layer and the intermediate layer is effective.

To evaluate our proposed method, we use the LibriSpeech dataset \cite{panayotov2015LibriSpeech} providing both large-scale acoustic and text training data.
Our experiments confirmed that CTC decoding accuracy was improved in student models applying interAED-KD.
Our best system showed similar or better word error rate (WER) than a conventional CTC-ASR system with LM beam search decoding while having a six times lower RTF.
Compared with conventional KD, the proposed method showed a WER reduction (WERR) of up to 7\%.

\section{CTC Acoustic Model}
\subsection{Connectionist Temporal Classification}
\label{sed:ctc}

In non-autoregressive ASR systems, the acoustic encoder \encam{} maps the input acoustic features $\acousticseq \in \rational^{\siglen \times \encdim}$ of length $\siglen$ and size $\encdim$ to the label sequence $\labelseq{} \in \vocab^{\labellen}$. The label sequence of length $\labellen$ contains tokens from the vocabulary $\vocab$ and has in most cases a different length than the acoustic features. To train such a system, the CTC loss function is used. The CTC loss maximises the log-likelihood over all possible alignment sequences $\alignseq \in \vocab^{\prime\siglen}$ between \acousticseq{} and \labelseq, where $\vocab^{\prime}$ is the vocabulary extended by a special blank symbol $\vocab^{\prime} \in \vocab \cup \{\blank\}$.
\begin{equation}
    \label{eq:ctc-prob}\log P_{\ctc} = \log\sum_{\alignseq \in \ctcmap^{-1}(\labelseq)} P(\alignseq | \acousticseq)
\end{equation}
\begin{equation}
    \label{eq:ctc-loss}\lossctc = -\log P_{\ctc}
\end{equation}
$\ctcmap$ is the operation that removes blank labels, repeated tokens from \alignseq{} to map to the label sequence. The probability of the alignment sequence is calculated based on the assumption of conditional independence between all elements $\alignelem_\sigelem$ in the alignment sequence $\alignseq$
\begin{equation}
\label{eq:p-ctc}
    P(\alignseq | \acousticseq) = \prod_\sigelem P(\alignelem_\sigelem | \acousticseq) 
\end{equation}
For some experiments we used intermediate CTC (interCTC) \cite{lee2021intermediate}, in which case we define \EqRef{eq:ctc-loss} as a weighted sum of both the CTC loss and the intermediate CTC loss (similar to the proposed training objective in \cite{lee2021intermediate}).

\subsection{Conformer Acoustic Model}
\label{sec:conformer}
To calculate \eqref{eq:p-ctc}, we use an acoustic encoder \encam{} based on the Conformer architecture \cite{gulati2020conformer}.
\encam{} consists of \enclayers{} encoder layers.
Each encoder layer $\enclayfun_\enclayer$ in the Conformer consists of a feed-forward network (\FFWD), a multi-head attention block, a convolutional layer, and another feed-forward network.
This macaron structure is followed by a layer normalisation block.
$\enclayfun_\enclayer$ transforms the input features $\feature_{\enclayer - 1} \in \rational^{\siglen \times \encdim}$ from its previous layer $\enclayer - 1$ to the output features $\feature_{\enclayer} \in \rational^{\siglen \times \encdim}$ by the following set of equations.
\begin{align}
    \feature^{\FFWD}_{\enclayer} &= \frac{1}{2}\ffwd{\feature_{\enclayer -1}} + \feature_{\enclayer - 1} \\
    \label{eq:conformer-mha}\feature^{\mha}_{\enclayer} &= \selfatt{\feature^{\FFWD}_{\enclayer}} + \feature^{\FFWD}_{\enclayer} \\
    \label{eq:conformer-conv}\feature^{\CONV}_{\enclayer} &= \conv\left(\feature_{\enclayer}^{\mha}\right) + \feature_{\enclayer}^{\mha} \\
    \feature_{\enclayer} &= \layernorm{\frac{1}{2}\ffwd{\feature^{\CONV}_{\enclayer}} + \feature^{\CONV}_{\enclayer}}
\end{align}
The probability of each element in the alignment sequence $\alignelem_{\sigelem}$ is calculated from the output of the final $\enclayfun_\enclayers$, followed by a linear transformation, and the softmax function.
\begin{equation}
    \mathbf{A} = \softmax{\ffwd{\enclayfun_\enclayers(\feature_{\enclayers - 1})}} 
\end{equation}
\begin{equation}
    P(\alignelem_{\sigelem} | \acousticseq) = \mathbf{A}[\sigelem]
\end{equation}

\section{External Language Model}
\label{sec:bert}
In this study, we follow the approach taken in previous research and use BERT as an external LM for KD.  BERT is a masked language model that predicts the probability of masked tokens $\word_\labelelem$ from all surrounding tokens $\wordseq_{\setminus \labelelem} = [\wordseq_{1:\labelelem-1}, \texttt{[MASK]}, \wordseq_{\labelelem+1:\labellen}]$. Doing so, the model relies on left and right context for its predictions. BERT uses the Transformer encoder architecture \cite{Vaswani2017attention}. A Transformer encoder $\enclm$ consists of several layers, where each layer comprises a multi-head attention block followed by a feed-forward layer. The probability of a masked token $\predword_{\labelelem}$ is calculated from the encoder output followed by a linear classifier and the softmax function.
\begin{equation}
    P(\predword_{\labelelem} | \wordseq_{\setminus \labelelem}) = \softmax{\ffwd{\enclm(\wordseq_{\setminus \labelelem})}}
\end{equation}
We obtain soft labels for KD from BERT by calculating BERT's token probabilities. To obtain the probabilities for the entire sequence $\predwordseq_\mathrm{BERT}$, we mask each token or word (multiple tokens) individually and concatenate the probabilities for all tokens $\predword_{\labelelem}, \labelelem \in \{1, \ldots, \labellen\}$ in the label sequence.

\section{Distillation into Intermediate Layers}
\label{sec:proposed_method}
\begin{figure}
    \centering
    \includegraphics[width=.95\columnwidth]{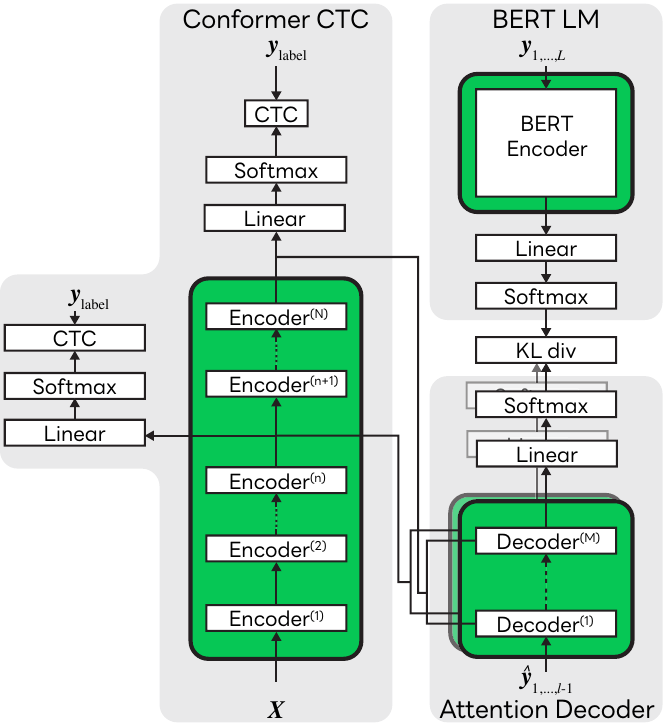}
    \caption{\label{fig:inter-decoder}Proposed model architecture with intermediate CTC loss and intermediate attention decoder for knowledge distillation. BERT's parameters are frozen during training the ASR model. The CTC decoders and the attention decoders share their parameters.}
\end{figure}

\begin{table*}
    \centering
    \caption{\label{tab:exp-results}WERs [\%] and RTFs on LibriSpeech 960 with greedy and LM beam search (BS) decoding strategies.}
    \begin{tabular}{ll|rrrr|rrrr|rrrr}
    \hline
    & & \multicolumn{4}{c|}{Greedy (RTF=0.003)} & \multicolumn{4}{c|}{BS (size=1, RTF=0.018)} & \multicolumn{4}{c}{BS (size=10, RTF=0.020)} \\
    & & \multicolumn{2}{c}{clean} & \multicolumn{2}{c|}{other} & \multicolumn{2}{c}{clean} & \multicolumn{2}{c|}{other} & \multicolumn{2}{c}{clean} & \multicolumn{2}{c}{other} \\
    Base model & KD method & dev & test & dev & test & dev & test & dev & test & dev & test & dev & test \\ \hline
    CTC & No KD & 3.46&3.63&8.34&8.45&3.09&3.27&7.56&7.67&2.53&2.79&6.50&6.56 \\
        & \textit{Conventional} AED-KD &2.95 & 3.17 & 7.13 & 7.26 & 2.62 & 2.82 & 6.45 & 6.58 & 2.29 & 2.42 & 5.75 & 5.89 \\
    \hspace{2mm}+interAED & No KD & 3.35 & 3.57 & 8.49 & 8.39 & 2.99 & 3.24 & 7.55 & 7.69 & 2.52 & 2.71 & 6.40 & 6.57 \\
        & \textit{Proposed} interAED-KD & 2.83 & 2.93 & 6.91 & 6.91 & 2.60 & 2.69 & 6.27 & 6.41 & 2.32 & 2.39 & 5.58 & 5.65 \\
    \hspace{4mm}+interCTC & No KD & 3.05 & 3.19 & 7.51 & 7.51 & 2.69 & 2.87 & 6.84 & 6.81 & 2.39 & 2.52 & 5.92 & 5.97 \\
        &\textit{Proposed} interAED-KD & \bf 2.60 & \bf 2.74 & \bf 5.95 & \bf 6.30 & \bf 2.35 & \bf 2.45 & \bf 5.48 & \bf 5.86 & \bf 2.16 & \bf 2.30 & \bf 4.93 & \bf 5.36 \\
    \hline    
    \end{tabular}
\end{table*}

In this section, we explain our proposed method for KD into intermediate encoder layers shown in \FigRef{fig:inter-decoder}.
In an autoregressive ASR system, an attention decoder $\attdec_\enclayers$ is used to predict the current token $\predword_\labelelem$.
This decoder may be a multi-layer Transformer decoder model.
For its prediction, $\attdec_\enclayers$ uses the output of $\enclayfun_\enclayers$ and the sequence of previously generated tokens $\predwordseq_{[1:\labelelem - 1]}= [\predword_1, \ldots, \predword_{\labelelem - 1}]$.
\begin{multline}
    \label{eq:att-dec-prob} P(\predword_{\labelelem} | \predwordseq_{[1:\labelelem - 1]}, \feature_{\enclayers}) = \\
    \softmax{\FFWD(\attdec_\enclayers(\predwordseq_{[1:\labelelem - 1]}, \feature_{\enclayers}))}
\end{multline}
Where $\FFWD$ is a classification layer that projects the decoder output to the vocabulary. $\attdec_\enclayers$ is conventionally trained with a cross entropy loss function, but to distil BERT's knowledge into the attention decoder, the Kullback-Leibler (KL) divergence is used. The KL divergence loss is calculated between the probability distribution of BERT's token predictions $\predword_{\mathrm{BERT},\labelelem}$ and the decoder's token predictions $\predword_{\mathrm{\attdec}_\enclayers,\labelelem}$.
\begin{equation}
    \label{eq:kl-div-loss}
    \losskl_{\enclayers} = 
    \sum_\labelelem P(\predword_{\mathrm{\attdec}_\enclayers,\labelelem}) \log(\frac{P(\predword_{\mathrm{\attdec}_\enclayers,\labelelem})}{P(\predword_{\mathrm{BERT},\labelelem})})
\end{equation}
As in \cite{futami2020distilling}, we use top-$K$ distillation for BERT's labels.
In top-$K$ distillation, we use only the probabilities for the top-$K$ predicted tokens instead of all tokens as distillation target.
In the experiments we set $K=10$.
For distillation into intermediate layers, we add further attention decoders $\attdec_\enclayer$ that use the outputs of intermediate encoder layers $\enclayfun_\enclayer$ \cite{komatsu2023interdecoder}.
These intermediate decoders $\attdec_\enclayer$ share their parameters with the final decoder $\attdec_\enclayers$ and are also trained to minimise the KL divergence with BERTs probability distribution.
\begin{equation}
    \label{eq:inter-kl-div-loss}
    \losskl_{\enclayer} =
    \sum_\labelelem P(\predword_{\mathrm{\attdec}_\enclayer,\labelelem}) \log(\frac{P(\predword_{\mathrm{\attdec}_\enclayer,\labelelem})}{P(\predword_{\mathrm{BERT},\labelelem})})
\end{equation}
Our distillation loss is a weighted sum of \EqRef{eq:kl-div-loss} and all $\kllosnum$ intermediate KL divergence losses.
\begin{equation}
    \label{eq:distill-loss}\lossdistill = (1 - \beta)\losskl_{\enclayers} +
    \beta\frac{1}{\kllosnum} \sum_{m=1}^{\kllosnum} \losskl_{\lfloor\frac{m\enclayers}{\kllosnum+1}\rfloor}
\end{equation}
This auxiliary loss from the AED is only used during training but the decoder is not required for CTC decoding.
That is, the final number of model parameters does not increase compared with a CTC only model.
The overall training loss is a weighted sum of \EqRef{eq:ctc-loss} and \EqRef{eq:distill-loss}.
\begin{equation}
    \label{eq:train-loss}\mathcal{L} = (1 - \alpha)\lossctc + \alpha\lossdistill
\end{equation}

\section{Experiments}
\label{sec:experiments}
\subsection{Dataset and Model Configuration}
\label{sec:exp-settings}
\begin{table}
    \centering
    \caption{\label{tab:model-param}Parameters for both ASR and LM encoders, and the auxiliary decoder.}
    \begin{tabular}{l|S[table-format=4.0]S[table-format=4.0]S[table-format=4.0]}
        \hline
        Model & {Conformer} & {BERT} & {Decoder} \\
        \hline
        Layers & 18 & 12 & 6\\
        Attention heads & & 8 & \\
        MHA dim & & 512 & \\
        Feed-forward dim & & 2048 & \\
        $|\vocab|$ & 5001 & 5005 & 5005 \\
        Total number of parameters & 124M & 41M & 30M \\
        \hline
    \end{tabular}
\end{table}

We conducted our experiments on the LibriSpeech \cite{panayotov2015LibriSpeech} dataset.
The dataset includes a normalised LM training dataset with 800M words that we used to train a BERT model and a 6-gram token LM.
We followed the ESPnet \cite{watanabe2018espnet} recipe for the ASR training data preparation.
To train the Conformer, we employed speed perturbation and SpecAugment \cite{park2019specaugment} to the 960h of acoustic training data.

\TabRef{tab:model-param} provides a summary of all models' parameters.
All models share the same base tokeniser with 5000 sentencepiece \cite{kudo-richardson-2018-sentencepiece} tokens.
The CTC decoder has an extra blank symbol, and BERT as well as the attention decoder have special tokens (\texttt{[CLS]}, \texttt{[SEP]}, etc.) added to their vocabulary.
Our attention decoder was a Transformer decoder with six layers and otherwise the same specifications as the encoders.
We applied intermediate AED and intermediate CTC at the 9th encoder layer.

For training and implementation of our LM and ASR models, we used Nvidia’s NeMo toolkit \cite{kuchaiev2019nemo}.
The ASR models were trained on 8$\times$ Nvidia V100 for 100 epochs with adaptive batch sizes ranging from 32 to 128, according to the audio length.
The learning rate was 1.4 and we used 10K warm up steps.
We set $\alpha$ in \EqRef{eq:train-loss} to 0.7 and $\beta$ in \EqRef{eq:distill-loss} to 0.5.

\subsection{Greedy Decoding}
\label{sec:ex-recog-result}
\TabRef{tab:exp-results} summarises the recognition results for LibriSpeech's clean and other subsets.
All models using KD show a lower WER than the base CTC-ASR model.
Our proposed KD on the intermediate encoder layers (interAED-KD) improved on KD only on the final layer (AED-KD), obtaining a relative WER reduction (WERR) of 7.6\% on test clean and 4.8\% on test other.
From this result, we see that the intermediate loss is effective in transferring more language information into the encoder compared to LM distillation only on the encoder output.

Combining interAED-KD and interCTC (interCTC-interAED-KD) resulted in the overall best model, achieving a WERR of 24.5\% on test clean and 25.4\% on test other compared with the base CTC-ASR model.
This finding suggests that both the interAED-KD and the interCTC objective functions help the acoustic encoder to learn complementary information.

As mentioned in \SecRef{sec:proposed_method}, we used Top-10 distillation in the experiments.
We experimented with increasing this number but did not notice any significant improvement over top-10 distillation.
Surprisingly, using BERT's hidden state representations instead of token probabilities for distillation resulted in worse results in our experiments.

Without further showing the results, we conducted additional experiments where we increased the number of intermediate losses for both interCTC and interAED-KD, but we did not observe any improvement on the LibriSpeech dataset.
This result aligns with similar findings in \cite{lee2021intermediate}.

\begin{table*}
    \centering
    \caption{\label{tab:decoding-comparison} Greedy decoding samples from LibriSpeech's test other subset with corresponding reference transcription.}
    \begin{tabular}{l|l}
    \hline
    Reference & i should have thought of it again when i was less busy may i go with you now \\
    CTC & i should have thought of it again when i was less busy may \textbf{ill} go with you now \\
    interCTC+interAED-KD CTC & i should have thought of it again when i was less busy may i go with you now \\
    interCTC+interAED-KD AED & i should have thought of it \textbf{but now that} i was \textbf{so tired} may i go with you now \\
    BERT & i should have thought of \textbf{you myself but} i was \textbf{so afraid but} i \textbf{speak} with you now \\
    \hline
    Reference & i don't believe all i hear no not by a big deal \\
    CTC & i \textbf{doanlie} all i hear no not by a big deal \\
    interCTC+interAED-KD CTC & i don't believe all i hear no not by a big deal \\
    interCTC+interAED-KD AED & i \textbf{do not} believe \textbf{what} i \textbf{say but} no not \textbf{quite} a \textbf{great} deal \\
    BERT & i don't believe \textbf{what} i \textbf{say} no not \textbf{for} a \textbf{great man} \\
    \hline
    \end{tabular}
\end{table*}

\subsection{LM Beam Search Decoding}
\label{sec:ex-assesing-kd}

In this section, we discuss the decoding results with LM beam search decoding. We performed beam search decoding with beam sizes of 1 and 10. LM weight and word insertion penalty were chosen from a rough parameter search on the dev clean and dev other set, respectively. The decoding results along with the RTFs are summarised in \TabRef{tab:exp-results}.

Beam search decoding with the Ngram LM and beam size 1 allows us to compare the contribution of both the KD and the Ngram LM. The CTC-ASR baseline achieved a worse WER than the models trained with our proposed interAED-KD and greedy decoding. When further increasing the beam size to 10, the CTC-ASR baseline's WER was lower than CTC+interAED-KD with greedy decoding. However, when comparing the RTFs of greedy decoding and beam search decoding, we found that greedy decoding has an at least six times lower RTF. These findings highlight the strengths of our proposed method, as it allows to keep the RTF of greedy, parallel decoding while giving better accuracy than autoregressive beam search decoding with an external LM.

Overall, beam search decoding significantly improved on greedy CTC decoding, even for the models trained with KD. However, this performance increase comes at the cost of longer decoding times and worse RTF, as \TabRef{tab:exp-results} shows. The RTF with beam search is over six times higher than with CTC greedy decoding. As comparison, when decoding with the AED of our model, the RTF was at 0.013 approximately four times higher than with greedy CTC decoding.

\subsection{Improvement by Intermediate KD}
Since our proposed method always combines intermediate AED \cite{komatsu2023interdecoder} with KD, we need to evaluate the contribution of KD separately from the intermediate AED. The results in \TabRef{tab:exp-results} clearly indicate the contribution of KD on improving the model parameters. When looking at the effect of interAED-KD itself, the language model information helped to reduce WER by 15-19\%. When using both interCTC and interAED, we saw a WERR of 14-20\% at the models that used KD.

These results show that the model could not learn significant language knowledge from the available amount of reference transcriptions. Using interAED-KD, BERT's language statistics could successfully be transferred into the acoustic encoder via an auxiliary AED and the AM could access this information without an external LM during decoding.

\subsection{Decoding Samples}
\label{sec:decoding-samples}

In our experiments, we compared the decoded outputs of different models. \TabRef{tab:decoding-comparison} shows some samples from LibriSpeech's test other subset and the corresponding reference transcriptions.

First, in the provided samples, the CTC-ASR baseline model already makes very few mistakes, but it can still output nonsensical tokens for unseen inputs as seen in the second example. Our proposed method reduces such mistakes significantly and in the shown examples outputs the correct words.

Second, when decoding the AED of our proposed method that was only trained with KL divergence but not cross entropy loss and the correct transcriptions, there are as expected significantly more errors in the decoded sequences. When we also decode the token probabilities from BERT and compare both the AED's and BERT's decoded sequences, we can see that they can exhibit large similarities. However, the examples also show that BERT's token probabilities are not always a good teacher signal and can itself be nonsensical at times.

\subsection{Discussion}

As seen \SecRef{sec:decoding-samples}, we discovered some limitations in the output probability distribution of BERT and room for further improvement of our proposed method. While BERT can predict words with similar semantic meaning as the masked word in utterances with long left and right context, for short utterances, BERT predicts words that are frequently seen in the training data. These words, while very likely in the context, can be very different from the original transcription.

Furthermore, we used LibriSpeech's normalised LM training data consisting of deduplicated sentences in alphabetical order to train BERT. While these data are suitable to train an Ngram LM, these data are not suitable for training neural network LMs like BERT because BERT falls short of learning important long-term context. We can expect to train a better BERT model from LibriSpeech's original unprocessed text corpus.

\section{Conclusion and Outlook}
\label{sec:conclusion_outlook}
We proposed a novel method for transferring external language model information into an AM by KD to improve the accuracy of a CTC-ASR model. It allows the AM to consider language knowledge when transcribing speech while maintaining the high speed of parallel CTC greedy decoding without using shallow fusion with an external LM. In addition, the proposed method uses intermediate AEDs only during training for KD, so the final model does not increase in its parameter size. However, when combined with an external LM and beam search during decoding even further WERR can be achieved. This also highlights opportunities for further improvements to our KD method. We are aiming to refine our method such that incorporating external LM information into ASR models for decoding only yields marginal to zero gains, and LM KD during model training is sufficient for optimal performance.

\bibliographystyle{IEEEbib}
\bibliography{mybib}

\end{document}

%% file: macros.tex
\def\ve#1{{\mathchoice{\mbox{\boldmath$\displaystyle #1$}}%
              {\mbox{\boldmath$\textstyle #1$}}%
              {\mbox{\boldmath$\scriptstyle #1$}}%
              {\mbox{\boldmath$\scriptscriptstyle #1$}}}}

\newcommand{\FigRef}[1]{Figure~\ref{#1}}	% Abbildung referenzieren
\newcommand{\TabRef}[1]{Table~\ref{#1}} 	% Tabelle referenzieren
\newcommand{\SecRef}[1]{Section~\ref{#1}} 	% Unterkapitel referenzieren
\newcommand{\EqRef}[1]{\eqref{#1}}	% Gleichung referenzieren (Befehl aus AMS package)

\def\ctc{\ensuremath{\mathrm{CTC}}}			% CTC
\def\mha{\ensuremath{\mathrm{MHA}}}         % multi-head attention
\def\CONV{\ensuremath{\mathrm{CONV}}}       % convolution
\def\FFWD{\ensuremath{\mathrm{FFWD}}}       % feed-forward

\def\rational{\ensuremath{\mathbb{R}}}      % rational numbers
\def\siglen{\ensuremath{S}}                 % input signal length
\def\sigelem{\ensuremath{s}}                % one index of the input signal
\def\labellen{\ensuremath{L}}               % label sequence length
\def\labelelem{\ensuremath{l}}              % index of the label sequence
\def\encdim{\ensuremath{D_\mathrm{enc}}}    % encoder hidden dimension
\def\enclayer{\ensuremath{n}}               % encoder layer
\def\enclayers{\ensuremath{N}}              % total number of encoder layer
\def\acousticseq{\ensuremath{\ve{X}}} % acoustic input sequence
\def\feature{\ensuremath{\ve{X}}}          % acoustic feature
\def\alignseq{\ensuremath{\ve{a}}}          % (valid) acoustic alignment sequence
\def\alignelem{\ensuremath{a}}              % element in the alignment sequence
\def\labelseq{\ensuremath{\ve{y}}}          % label sequence
\def\vocab{\ensuremath{\mathcal{U}}}		% Vocabulary
\def\encam{\ensuremath{\mathrm{Enc_{AM}}}}           % acoustic encoder
\def\enclm{\ensuremath{\mathrm{Enc_{LM}}}}           % acoustic encoder
\def\enclayfun{\ensuremath{\mathrm{EncLayer}}}  % one layer of the encoder
\def\blank{\ensuremath{\mathcal{\epsilon}}} % CTC blank symbol
\def\word{\ensuremath{y}}                   % word
\def\wordseq{\ensuremath{\ve{y}}}           % word sequence
\def\predword{\ensuremath{\hat{y}}}         % predicted word
\def\predwordseq{\ensuremath{\ve{\hat{y}}}} % predicted word sequence
\def\attdec{\ensuremath{\mathrm{Dec}}}       % attention decoder
\def\kllosnum{\ensuremath{M}}               % total number of KL div losses
\def\lossdistill{\ensuremath{\mathcal{L}^\mathrm{distill}}}  % distillation loss
\def\lossctc{\ensuremath{\mathcal{L}^\mathrm{CTC}}}  % CTC loss
\def\losskl{\ensuremath{\mathcal{L}^\mathrm{KL}}}  % KL div loss

\DeclareMathOperator{\ctcmap}{\ensuremath{\mathrm{\psi}}}
\newcommand{\softmax}[1]{\ensuremath{\mathrm{Softmax}\left(#1\right)}}    % softmax function
\newcommand{\selfatt}[1]{\ensuremath{\mathrm{SelfAttention}\left(#1\right)}}   % self-attention
\newcommand{\ffwd}[1]{\ensuremath{\mathrm{FFWD}\left(#1\right)}}   % feed-forward layer
\DeclareMathOperator{\conv}{\ensuremath{\mathrm{Convolution}}}   % convolution layer
\newcommand{\layernorm}[1]{\ensuremath{\mathrm{LayerNorm}\left(#1\right)}}   % layer norm

%% file: main.bbl
\begin{thebibliography}{10}

\bibitem{meng2021internal}
Zhong Meng, Sarangarajan Parthasarathy, Eric Sun, Yashesh Gaur, Naoyuki Kanda,
  Liang Lu, Xie Chen, Rui Zhao, Jinyu Li, and Yifan Gong,
\newblock ``Internal language model estimation for domain-adaptive end-to-end
  speech recognition,''
\newblock in {\em 2021 IEEE Spoken Language Technology Workshop (SLT)}, 2021,
  pp. 243--250.

\bibitem{meng2021internaltraining}
Zhong Meng, Naoyuki Kanda, Yashesh Gaur, Sarangarajan Parthasarathy, Eric Sun,
  Liang Lu, Xie Chen, Jinyu Li, and Yifan Gong,
\newblock ``Internal language model training for domain-adaptive end-to-end
  speech recognition,''
\newblock in {\em 2021 IEEE International Conference on Acoustics, Speech and
  Signal Processing (ICASSP)}, 2021, pp. 7338--7342.

\bibitem{Chorowski2017}
Jan Chorowski and Navdeep Jaitly,
\newblock ``{Towards Better Decoding and Language Model Integration in Sequence
  to Sequence Models},''
\newblock in {\em Proc. Interspeech 2017}, 2017, pp. 523--527.

\bibitem{kannan18_icassp}
Anjuli Kannan, Yonghui Wu, Patrick Nguyen, Tara~N. Sainath, ZhiJeng Chen, and
  Rohit Prabhavalkar,
\newblock ``{An Analysis of Incorporating an External Language Model into a
  Sequence-to-Sequence Model},''
\newblock in {\em 2018 IEEE International Conference on Acoustics, Speech and
  Signal Processing (ICASSP)}, 2018, pp. 1--5828.

\bibitem{Graves2012Sequence}
Alex Graves,
\newblock ``{S}equence {T}ransduction with {R}ecurrent {N}eural {N}etworks,''
\newblock in {\em International Conference of Machine Learning (ICML) Workshop
  on Representation Learning}, 2012.

\bibitem{graves2013speech}
Alex Graves, Abdel-rahman Mohamed, and Geoffrey Hinton,
\newblock ``Speech recognition with deep recurrent neural networks,''
\newblock in {\em 2013 IEEE International Conference on Acoustics, Speech and
  Signal Processing (ICASSP)}, 2013, pp. 6645--6649.

\bibitem{graves2006connectionist}
Alex Graves, Santiago Fern\'{a}ndez, Faustino Gomez, and J\"{u}rgen
  Schmidhuber,
\newblock ``{C}onnectionist {T}emporal {C}lassification: {L}abelling
  {U}nsegmented {S}equence {D}ata with {R}ecurrent {N}eural {N}etworks,''
\newblock in {\em Proceedings of the 23rd International Conference on Machine
  Learning}, 2006, p. 369–376.

\bibitem{hinton2015distilling}
Geoffrey Hinton, Oriol Vinyals, and Jeffrey Dean,
\newblock ``Distilling the {K}nowledge in a {N}eural {N}etwork,''
\newblock in {\em NIPS Deep Learning and Representation Learning Workshop},
  2015.

\bibitem{devlin2019bert}
Jacob Devlin, Ming-Wei Chang, Kenton Lee, and Kristina Toutanova,
\newblock ``{BERT}: Pre-training of {D}eep {B}idirectional {T}ransformers for
  {L}anguage {U}nderstanding,''
\newblock in {\em NAACL HLT}, 2019, pp. 4171--4186.

\bibitem{futami2020distilling}
Hayato Futami, Hirofumi Inaguma, Sei Ueno, Masato Mimura, Shinsuke Sakai, and
  Tatsuya Kawahara,
\newblock ``{Distilling the {K}nowledge of {BERT} for {S}equence-to-{S}equence
  {ASR}},''
\newblock in {\em INTERSPEECH}, 2020, pp. 3635--3639.

\bibitem{bai2021fast}
Ye~Bai, Jiangyan Yi, Jianhua Tao, Zhengkun Tian, Zhengqi Wen, and Shuai Zhang,
\newblock ``Fast {E}nd-to-{E}nd {S}peech {R}ecognition {V}ia
  {N}on-{A}utoregressive {M}odels and {C}ross-{M}odal {K}nowledge
  {T}ransferring {F}rom {BERT},''
\newblock {\em IEEE/ACM Transactions on Audio, Speech, and Language
  Processing}, vol. 29, pp. 1897--1911, 2021.

\bibitem{kubo2022knowledge}
Yotaro Kubo, Shigeki Karita, and Michiel Bacchiani,
\newblock ``Knowledge {T}ransfer from {L}arge-{S}cale {P}retrained {L}anguage
  {M}odels to {E}nd-{T}o-{E}nd {S}peech {R}ecognizers,''
\newblock in {\em 2022 IEEE International Conference on Acoustics, Speech and
  Signal Processing (ICASSP)}, 2022, pp. 8512--8516.

\bibitem{higuchi-etal-2022-bert}
Yosuke Higuchi, Brian Yan, Siddhant Arora, Tetsuji Ogawa, Tetsunori Kobayashi,
  and Shinji Watanabe,
\newblock ``{BERT} meets {CTC}: New formulation of end-to-end speech
  recognition with pre-trained masked language model,''
\newblock in {\em Findings of the Association for Computational Linguistics:
  EMNLP 2022}, 2022, pp. 5486--5503.

\bibitem{jiao-etal-2020-tinybert}
Xiaoqi Jiao, Yichun Yin, Lifeng Shang, Xin Jiang, Xiao Chen, Linlin Li, Fang
  Wang, and Qun Liu,
\newblock ``{T}iny{BERT}: Distilling {BERT} for natural language
  understanding,''
\newblock in {\em Findings of the Association for Computational Linguistics:
  EMNLP 2020}, 2020, pp. 4163--4174.

\bibitem{sun-etal-2020-mobilebert}
Zhiqing Sun, Hongkun Yu, Xiaodan Song, Renjie Liu, Yiming Yang, and Denny Zhou,
\newblock ``{M}obile{BERT}: a compact task-agnostic {BERT} for resource-limited
  devices,''
\newblock in {\em Proceedings of the 58th Annual Meeting of the Association for
  Computational Linguistics}, 2020, pp. 2158--2170.

\bibitem{hentschel2021making}
Michael Hentschel, Emiru Tsunoo, and Takao Okuda,
\newblock ``{Making Punctuation Restoration Robust and Fast with Multi-Task
  Learning and Knowledge Distillation},''
\newblock in {\em 2021 IEEE International Conference on Acoustics, Speech and
  Signal Processing (ICASSP)}, 2021, pp. 7773--7777.

\bibitem{yoon2022interkd}
Ji~Won Yoon, Beom~Jun Woo, Sunghwan Ahn, Hyeonseung Lee, and Nam~Soo Kim,
\newblock ``{Inter-KD: Intermediate Knowledge Distillation for CTC-Based
  Automatic Speech Recognition},''
\newblock in {\em 2022 IEEE Spoken Language Technology Workshop (SLT)}, 2023,
  pp. 280--286.

\bibitem{komatsu2023interdecoder}
Tatsuya Komatsu and Yusuke Fujita,
\newblock ``{Interdecoder: using Attention Decoders as Intermediate
  Regularization for CTC-Based Speech Recognition},''
\newblock in {\em 2022 IEEE Spoken Language Technology Workshop (SLT)}, 2023,
  pp. 46--51.

\bibitem{deng2022improving}
Keqi Deng, Zehui Yang, Shinji Watanabe, Yosuke Higuchi, Gaofeng Cheng, and
  Pengyuan Zhang,
\newblock ``Improving {N}on-{A}utoregressive {E}nd-to-{E}nd {S}peech
  {R}ecognition with {P}re-{T}rained {A}coustic and {L}anguage {M}odels,''
\newblock in {\em 2022 IEEE International Conference on Acoustics, Speech and
  Signal Processing (ICASSP)}, 2022, pp. 8522--8526.

\bibitem{panayotov2015LibriSpeech}
Vassil Panayotov, Guoguo Chen, Daniel Povey, and Sanjeev Khudanpur,
\newblock ``{Librispeech: An ASR corpus based on public domain audio books},''
\newblock in {\em 2015 IEEE International Conference on Acoustics, Speech and
  Signal Processing (ICASSP)}, 2015, pp. 5206--5210.

\bibitem{lee2021intermediate}
Jaesong Lee and Shinji Watanabe,
\newblock ``Intermediate {L}oss {R}egularization for {CTC}-{B}ased {S}peech
  {R}ecognition,''
\newblock in {\em 2021 IEEE International Conference on Acoustics, Speech and
  Signal Processing (ICASSP)}, 2021, pp. 6224--6228.

\bibitem{gulati2020conformer}
Anmol Gulati, James Qin, Chung-Cheng Chiu, Niki Parmar, Yu~Zhang, Jiahui Yu,
  Wei Han, Shibo Wang, Zhengdong Zhang, Yonghui Wu, and Ruoming Pang,
\newblock ``{Conformer: {C}onvolution-augmented {T}ransformer for {S}peech
  {R}ecognition},''
\newblock in {\em INTERSPEECH}, 2020, pp. 5036--5040.

\bibitem{Vaswani2017attention}
Ashish Vaswani, Noam Shazeer, Niki Parmar, Jakob Uszkoreit, Llion Jones,
  Aidan~N Gomez, \L~ukasz Kaiser, and Illia Polosukhin,
\newblock ``Attention is {A}ll you {N}eed,''
\newblock in {\em Advances in Neural Information Processing Systems}. 2017,
  vol.~30, Curran Associates, Inc.

\bibitem{watanabe2018espnet}
Shinji Watanabe, Takaaki Hori, Shigeki Karita, Tomoki Hayashi, Jiro Nishitoba,
  Yuya Unno, Nelson {Enrique Yalta Soplin}, Jahn Heymann, Matthew Wiesner,
  Nanxin Chen, Adithya Renduchintala, and Tsubasa Ochiai,
\newblock ``{ESPnet: End-to-End Speech Processing Toolkit},''
\newblock in {\em INTERSPEECH}, 2018, pp. 2207--2211.

\bibitem{park2019specaugment}
Daniel~S. Park, William Chan, Yu~Zhang, Chung-Cheng Chiu, Barret Zoph, Ekin~D.
  Cubuk, and Quoc~V. Le,
\newblock ``{SpecAugment: A Simple Data Augmentation Method for Automatic
  Speech Recognition},''
\newblock in {\em INTERSPEECH}, 2019, pp. 2613--2617.

\bibitem{kudo-richardson-2018-sentencepiece}
Taku Kudo and John Richardson,
\newblock ``{S}entence{P}iece: A simple and language independent subword
  tokenizer and detokenizer for neural text processing,''
\newblock in {\em EMNLP}, 2018, pp. 66--71.

\bibitem{kuchaiev2019nemo}
Oleksii Kuchaiev, Jason Li, Huyen Nguyen, Oleksii Hrinchuk, Ryan Leary, Boris
  Ginsburg, Samuel Kriman, Stanislav Beliaev, Vitaly Lavrukhin, Jack Cook,
  et~al.,
\newblock ``Nemo: a toolkit for building ai applications using neural
  modules,''
\newblock {\em arXiv preprint arXiv:1909.09577}, 2019.

\end{thebibliography}
